\documentclass[10pt,twocolumn,letterpaper]{article}

\usepackage[pagenumbers]{cvpr} 

\usepackage{booktabs}
\usepackage{multirow}
\usepackage{overpic}
\usepackage{pifont}
\usepackage{graphicx}
\usepackage[table]{xcolor}
\usepackage{listings}
\usepackage{tcolorbox}

\newcommand{\model}{VACoT}
\newcommand{\bench}{AdvOCR}

\newcommand{\myparagraph}[1]{\vspace{1mm}\noindent{{\textbf{\textit{#1}}}}}
\newcommand{\cmark}{\ding{51}}
\newcommand{\xmark}{\ding{55}}

\definecolor{ao(english)}{rgb}{0.0, 0.5, 0.0}
\definecolor{deepred}{RGB}{177, 29, 5}

\definecolor{darkNavy}{RGB}{34, 36, 73}
\definecolor{lightBlue}{RGB}{156, 179, 212}
\definecolor{dustyRose}{RGB}{175, 90, 118}
\tcbuselibrary{listings}
\tcbuselibrary{breakable}

\definecolor{cvprblue}{rgb}{0.21,0.49,0.74}
\usepackage[pagebackref,breaklinks,colorlinks,allcolors=cvprblue]{hyperref}

\def\paperID{} 
\def\confName{CVPR}
\def\confYear{2026}

\title{VACoT: Rethinking Visual Data Augmentation with VLMs}

\author{Zhengzhuo Xu$^{1,2}$ ~~ Chong Sun$^{2}$ ~~ SiNan Du$^{1}$ ~~ Chen Li$^{2}$ ~~ Jing LYU$^2$  ~~ Chun Yuan$^{1}$ \\
$^1$Tsinghua University ~~ $^2$WeChat Vision, Tencent Inc.
}

\begin{document}
\maketitle
\begin{abstract}
    While visual data augmentation remains a cornerstone for training robust vision models, it has received limited attention in visual language models (VLMs), which predominantly rely on large-scale real data acquisition or synthetic diversity. Consequently, they may struggle with basic perception tasks that conventional models handle reliably. Given the substantial cost of pre-training and fine-tuning VLMs, continue training on augmented data yields limited and diminishing returns. In this paper, we present \textbf{V}isual \textbf{A}ugmentation \textbf{C}hain-\textbf{o}f-\textbf{T}hought (\textbf{VACoT}), a framework that dynamically invokes image augmentations during model inference. By incorporating post-hoc transformations such as denoising, VACoT substantially improves robustness on challenging and out-of-distribution inputs, especially in OCR-related adversarial scenarios. Distinct from prior approaches limited to local cropping, VACoT integrates a structured collection of general visual augmentations, broadening the query image views while reducing training complexity and computational overhead with efficient agentic reinforcement learning. We propose a conditional reward scheme that encourages necessary augmentation while penalizing verbose responses, ensuring concise and effective reasoning in perception tasks. We demonstrate the superiority of VACoT with extensive experiments on 13 perception benchmarks and further introduce AdvOCR to highlight the generalization benefits of post-hoc visual augmentations in adversarial scenarios.
\end{abstract}
    
\section{Introduction}
\label{sec_intro}

\begin{figure}[t]
\centering
  \begin{overpic}[width=\linewidth, grid=False]{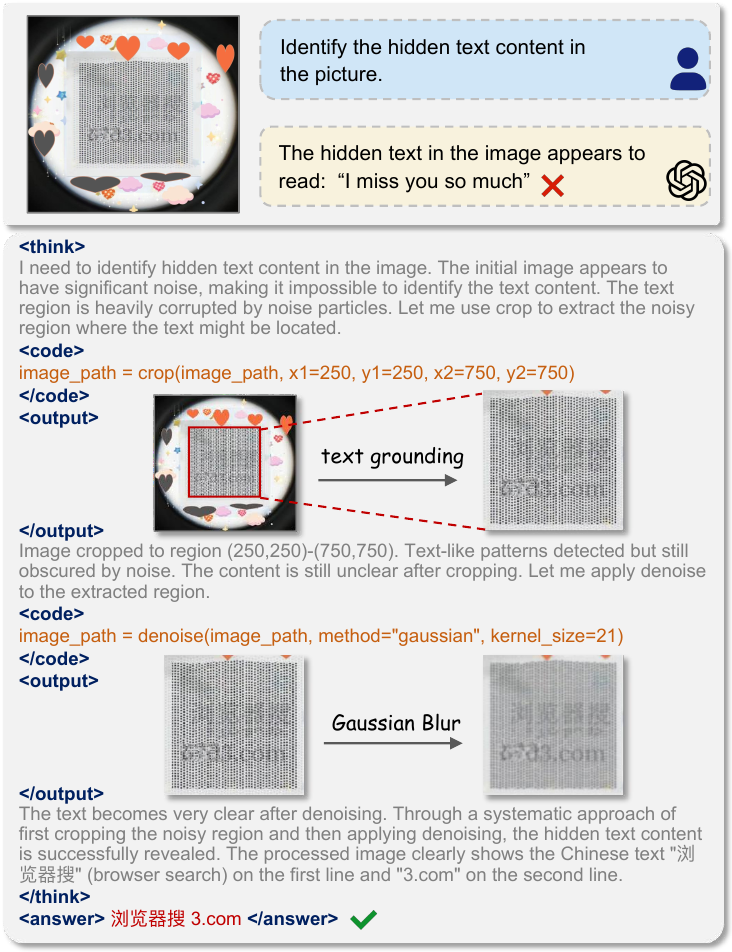}
  \end{overpic}
  \captionsetup{skip=3pt}      
  \caption{VACoT inference example. We address adversarial text recognition through iterative post-hoc visual augmentations.}
\vspace{-20pt}
\label{fig_teaser}
\end{figure}

With scaling laws empirically validated on Large Language Models (LLMs), Visual Language Models (VLMs) have also achieved breakthroughs in image perception~\cite{Qwen2, gemini2.5}, reasoning~\cite{R1Survey, gpt}, and generation~\cite{seedance, bagel} tasks. While VLMs outperform traditional deep models on tasks such as key information extraction~\cite{survey_kie} and security auditing~\cite{survey_ocr}, they are highly vulnerable to adversarial samples. Fig.~\ref{fig_teaser} shows that slight perturbations can mislead VLM perception results, enabling harmful content to evade system detection. Hence, ensuring adversarial robustness of VLMs remains an open and challenging problem.

Previous models rely on extensive visual data augmentations~\cite{augmix, survey_augment, Sptsv2} to improve perception robustness. However, such augmentation introduces limited additional multimodal knowledge while substantially increasing computational costs during Pre-Training (PT) and Supervised Fine-Tuning (SFT). Hence, existing VLMs prioritizes large-scale real-world data or diverse domain-specific synthetic data, instead of more augmented visual questions. Recent studies~\cite{deepeyes, Thyme, Vlrethinker, simpleo3, perceptionr1} have explored the \textit{thinking with bounding boxes} paradigm, which employs local cropping to focus model attention on specific regions. However, this strategy constitutes merely a specific instance of visual information filtering. More broadly, we advocate for \textit{thinking with augmentation}, reformulating cropping and other visual augmentations as post-hoc processing steps. This enhances model perceptual robustness while avoiding the training overhead caused by redundant image augmentations. As illustrated in Fig.~\ref{fig_teaser}, we propose Visual Augmentation CoT (VACoT) based on chat history concatenation. It stops autoregressive generation at designated tokens to either produce visual augmentations or terminate responses. By re-integrating returned information or augmented images into the conversation, we adopt end-to-end optimization via agentic Reinforcement Learning (RL). We wrap all augmentations into lightweight API calls to reduce the training difficulty and avoid uncontrollable actions. 

Existing instruction-based agents struggle to comprehend when and what to apply visual augmentations based solely on textual descriptions. Hence, VACoT adopts a three-stage training pipeline to overcome it. \textit{Stage} 1: We employ knowledge SFT with a difficulty-based data filtering strategy to efficiently enhance the model's foundational capabilities. \textit{Stage} 2: We employ format SFT for visual augmentation cold-start initialization. Generating reliable trajectory data for post-hoc augmentation requires substantial manual effort and cost. Hence, we prompt the teacher models to deliberately insert random API calls when rewriting answers. This stage focuses on learning the correct calling format, even if these calls are not semantically relevant to the current query. \textit{Stage} 3: We perform end-to-end agentic reinforcement learning with carefully designed reward signals, enabling the model to adaptively determine when and which augmentations to apply. We employ Qwen3~\cite{Qwen3} as a teacher model to provide reward signals for verifying both answer correctness and formatting consistency.
We penalize unnecessarily long reasoning traces that contribute limited performance gains. To this end, we incorporate the consistency reward~\cite{r1reward} to assess the reasoning process, and further introduce a novel conditional API-call reward that promotes effective visual augmentation while preventing sequence explosion caused by indiscriminate trials.

Extensive evaluations on 13 public benchmarks demonstrate that our post-hoc visual augmentation significantly enhances perceptual capabilities, particularly on tasks requiring fine-grained recognition or robustness against adversarial text. We further introduce {\bench}, a challenging benchmark comprising 100 adversarial samples for perception evaluation. Although it focuses solely on perception tasks, the benchmark is particularly challenging because of the adversarially modified images. Our {\model} could respond directly to clear queries and leverage iterative augmentations on adversarial samples to achieve high-confidence answers. In summary, our key contributions are:
\setlength{\leftmargini}{20pt}
\begin{enumerate}[itemsep=1pt, topsep=1pt, label=\alph*)]
    \item We propose {\model} that integrates post-hoc visual augmentations into agentic workflows, significantly enhancing VLM robustness against adversarial inputs.
    \item We devise an efficient three-stage training pipeline to learn adaptive visual augmentation strategies without prohibitive manual annotation.
    \item We design a conditional API-call reward to effectively promote necessary augmentations while penalizing redundant actions in reinforcement learning.
    \item We present a challenging benchmark called {\bench} to rigorously evaluate the perceptual robustness of VLMs with fine-grained or adversarial images.
\end{enumerate}

\section{Related Works}
\label{sec_related}

\myparagraph{Vision Language Models.}
Research has advanced from sophisticated projectors like Q-Formers~\cite{BLIP2, Flamingo} for modality alignment to more streamlined designs utilizing linear projectors coupled with instruction tuning~\cite{llava, BLIP3}. For model ecosystems, proprietary models like the GPT~\cite{gpt} and Claude~\cite{Claude} series represent the state-of-the-art in performance. Meanwhile, the open-source community has been significantly advanced by series such as LLaMA~\cite{llama, llama-2} and its derivatives (e.g., LLaVA~\cite{llava,llava15}), with recent series like QwenVL~\cite{QwenVL, Qwen2VL, Qwen2.5VL} and InternVL~\cite{internlm-xcomposer, internlm-xcomposerv2, internlm2, internvl3} pushing the performance frontier. The DeepSeek~\cite{deepseekv2, deepseekvl2} and Mistral~\cite{mistral7b} series have further contributed by exploring scalable architectures like mixture of experts.

\myparagraph{Visual Reasoning.}
CoT~\cite{CoT} and its variants~\cite{CCoT, MCoT} enhance reasoning by decomposing complex problems into sequential steps. This paradigm has been significantly advanced by models like GPT-o1~\cite{gpt} and DeepSeek-R1~\cite{DeepseekR1}, which employ reinforcement learning to bolster reasoning capabilities. Performance has also been improved by enabling longer output sequences directly~\cite{eureka, lmmr1, Reasonrft, R1Survey}. Subsequent efforts have explored multimodal CoT, where the reasoning structure is dynamically refined across multiple generation turns by incorporating region cropping~\cite{MCoT} or external knowledge~\cite{MVoT} to bolster interaction with visual tokens. Recent work~\cite{perceptionr1, Vlrethinker, deepeyes, openvisionreasoner, Thyme} focuses on end-to-end training for tool invocation, aiming to equip models with autonomous multimodal reasoning. Nevertheless, existing methods are still confined to local region positioning for addressing fine-grained recognition.

\begin{figure*}[ht!]
\centering
    \includegraphics[width=1\linewidth]{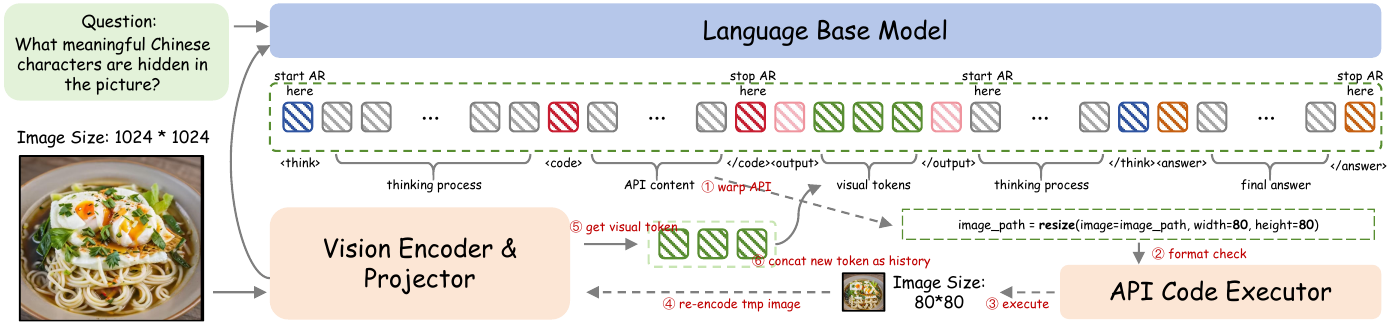}
    \captionsetup{skip=3pt}    
    \caption{The overall architecture of {\model}. We leverage stop-words to achieve iterative post-hoc visual augmentation, providing more diverse image perspectives and higher-quality visual interactions compared to cropping-only agentic models.}
\vspace{-15pt}
\label{fig_arch}
\end{figure*}

\myparagraph{Perception Benchmarks.}
Existing perception benchmarks primarily focus on chart understanding~\cite{chartqa, chartqapro, charxiv, ChartBench} and document recognition tasks~\cite{AI2D, TextVQA, infovqa, Docvqa, OCRBench}. However, state-of-the-art models have largely saturated these benchmarks, achieving performance exceeding 90\%. While recent benchmarks~\cite{ocrbenchv2, zerobench} have increasingly emphasized reasoning complexity, the perceptual aspect of textual content in images remains relatively basic. In contrast, {\bench} is derived from real-world failures and is designed to assess robust text perception in adversarial scenarios.

\section{Method}

\subsection{{\model}}

\myparagraph{Notation.}
Let $I^{0}$ be the query image and encoder $E(\cdot)$ outputs visual tokens $v^{k} = E(I^{k})$ for the $k$-th state. Base LLM $L(\cdot)$ generates token sequence $y_{1:t}$ up to $t$. Available visual augmentation set is $\mathcal{A}$ and state $k$ action is $a^{k} \in \mathcal{A} \cup \{\varnothing\}$. $\mathrm{Exec}(I, a)$ applies $a$ to $I$ to return the transformed image or error message $e$. $H_t$ is the full context at step $t$, and $\kappa$ is the generate stop token set.

\myparagraph{Visual Augmentation CoT.} Fig.~\ref{fig_arch} shows the architecture of {\model}, which is based on Qwen2.5VL-3B~\cite{Qwen2.5VL}. At reasoning step $t^k$, the language model $L(\cdot)$ autoregressively generates tokens conditioned on the current context $H_{t-1}$:
\begin{equation}
   P(y_t \mid H_{t-1}) = L(y_t \mid H_{t-1}). 
\end{equation}

When the model emits token $y^k_{t}$ in $\kappa$, the token sequence is parsed into a candidate operation $\hat{a}^{k}$ via a syntax parser $\hat{a}^{k} = \texttt{parse}(y_{t^{k-1}}:y_{t^k})$, where $\hat{a}^{k}$ represents the span of tokens enclosed by the \texttt{<code></code>} tag. The selected operation is executed through the API code executor:
\begin{equation}
e^{k}, I^{k} = \mathrm{Exec}(I^{k-1}, a^{k}).
\end{equation}

The augmented image $I^{k}$ is re-encoded by the vision encoder $E(\cdot)$ to obtain new visual tokens:
\begin{equation}
h^{k}  = \begin{cases} 
v^{k} = E(I^{k}), & \text{if } I^{k} \text{ is available} \\
e^{k}, & \text{otherwise}
\end{cases}
\end{equation}

The chat history is updated with new visual embedding $v^{k}$ or execution error message $e^{k}$ as follows:
\begin{equation}
H_{t} = H_{t-1} \oplus \texttt{<output>} h^{k} \texttt{</output>},
\end{equation}
where $\oplus$ denotes message concatenation. This closed-loop process enables iterative token generation, visual augmentation, and feedback integration into subsequent steps.

\myparagraph{Augmentation Design.} We first implement several augmentations as API calls (e.g., \texttt{brightness}, \texttt{contrast}, \texttt{saturation}, \texttt{sharpening}, and \texttt{thresholding}). However, manual evaluations reveal that these post-hoc operations have a negligible impact on the model’s perceptual capability. Consequently, we refine the set to more stable operations for $\mathcal{A}$: \texttt{crop}, \texttt{resize} ($\uparrow$/$\downarrow$), \texttt{rotate}, \texttt{flip}, \texttt{denoise} (filtering), and \texttt{edge} (edge detection). API calls are only executed if the operation and its parameters match. Otherwise, the corresponding error message is returned. The call success rate is ensured via supervised fine-tuning, reward design, and rule-based re-checks, which prevent token waste and uncontrolled code generation.

\subsection{Agentic RL}
\label{sec_agentic_rl}
\myparagraph{Strategy.}
We employ the on-policy GRPO algorithm~\cite{grpo} to guide baseline to effectively utilize visual augmentation tools. The core improvement lies in handling the trajectory token discrepancies introduced by multiple rounds of tool invocations. The $i$-th complete generation trajectory of {\model} is formulated as:
\begin{equation}
    s_i = \{(v^0, s_i^{Q}) \oplus \sum_{k}^{\oplus} (s_i^{k}, a_{i}^k, h_{i}^k) \oplus s_i^{A}\},
\end{equation}
where $s$ denotes the token sequence, with superscripts $Q$ and $A$ representing the initial query and the final answer, respectively. The operator $\sum^{\oplus}$ indicates the sequential concatenation of grouped elements. Accordingly, the token sequence used for loss computation is defined as:
\begin{equation}
    \tau_i = \{s_i^{Q} \oplus \sum_{k}^{\oplus} (s_i^{k}, a_{i}^k) \oplus s_i^{A}\}.
\end{equation}

Given a generation policy $\pi_{\theta}$ parameterized by $\theta$ and a rollout set $G$, the optimization objective is:
\begin{equation}
    \mathcal{J}(\theta) = 
    \frac{1}{\sum_{i=1}^G |\tau_i|}
    \sum_{i=1}^G \sum_{j=1}^{|\tau_i|}
    \sigma(r_{i,j}) - \beta \cdot \mathbb{D}_{\text{KL}}(\pi_\theta \| \pi_{\text{ref}}),
\end{equation}
where $r$ is trajectory reward, $\sigma(\cdot)$ is intra-trajectory normalization, and $\beta$ controls the KL-divergence regularization.

\myparagraph{Reward.} For each trajectory $s_i$, we define 5 rewards with reward model Qwen3-30B-A3B~\cite{Qwen3}:
\textbf{1)} $R_\text{vqa}$: We evaluate the correctness of the reasoning and final answer based on the last 500 characters of $s_i$. $R_\text{vqa}$ is continuous in [0, 1].
\textbf{2)} $R_\text{fmt}$: We use regex-based matching to detect \texttt{<think>} and \texttt{<answer>} tags~\cite{DeepseekR1}. $R_\text{fmt}$ is binary in 0 or 1.
\textbf{3)} $R_\text{cst}$: We evaluates whether the reasoning process contains redundant repetition and whether the final answer is logically consistent with the reasoning. $R_\text{cst}$ is continuous in [0, 1].
\textbf{4)} $R_\text{api}$: We detect all API calls in $s_i$ via regex and assess the validity of the operation names and parameters. $R_\text{api}$ is binary in 0 or 1.
\textbf{5)} $R_\text{suc}$: This reward is re-weighted by $R_\text{vqa}$, API calls times $k$ in $s_i$ and maximum allowed times $K$, which is:
\begin{equation}
R_\text{suc} =
\begin{cases}
0, & \text{if } R_{\text{vqa}} < 0.5, \\
1, & \text{if } R_{\text{vqa}} \ge 0.5, k \le 2, \\
1 - \dfrac{k - 2}{K - 2}, & \text{if } R_{\text{vqa}} \ge 0.5, k \in (2,K], \\
0, & \text{if } R_{\text{vqa}} \ge 0.5, k > K.
\end{cases}
\end{equation}

This reward discourage excessively long reasoning or frequent tool calls in perception-oriented tasks. The final reward is weighted ($\hat{R}$) as follows:
\begin{equation}
    r_i = \hat{R}_\text{vqa} + \hat{R}_\text{fmt} + \hat{R}_\text{cst} + \hat{R}_\text{api} + \hat{R}_\text{suc},
\end{equation}
where the weights we adopt are [1, 0.25, 0.5, 0.25, 0.5].

\subsection{Training.}
\label{sec_method_training}
\begin{figure}[ht]
  \centering
   \includegraphics[width=\linewidth]{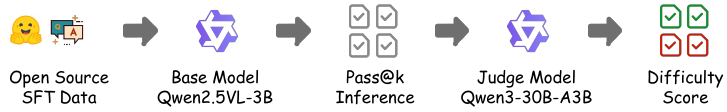}
   \captionsetup{skip=3pt}  
   \caption{Illustration of pass@k data filter. The difficulty score is aligned with the number of correct answers for pass@k inference.}
\vspace{-10pt}
\label{fig_sft_passk}
\end{figure}

\myparagraph{Stage 1: Knowledge Enhancement.}
Given that the baseline model has undergone extensive pre-training, we apply a difficulty-based data filtering strategy to improve knowledge SFT efficiency. First, we collect approximately 4M high-quality open-source QA pairs~\cite{cosyn, AI2D, chartqa, ArxivQA, Docvqa}. As illustrated in Fig.~\ref{fig_sft_passk}, we perform \textit{pass@4} inference with the Qwen2.5-VL-3B model using vLLM~\cite{vllm} and high generative diversity. Qwen3-30B-A3B~\cite{Qwen3} evaluates each answer to assign the difficulty score according to the number of error answer. We randomly select 10\% of the difficulty level 0 samples while retaining all samples with difficulty levels between 1 and 3. For samples rated at difficulty level 4, we employ Qwen2.5-VL-72B~\cite{Qwen2.5VL} to verify the semantic validity of each QA pair and discard those deemed inherently unanswerable. After filtering, we obtain a total of 411K samples for knowledge-enhanced SFT.

\begin{figure}[ht]
  \centering
   \includegraphics[width=0.9\linewidth]{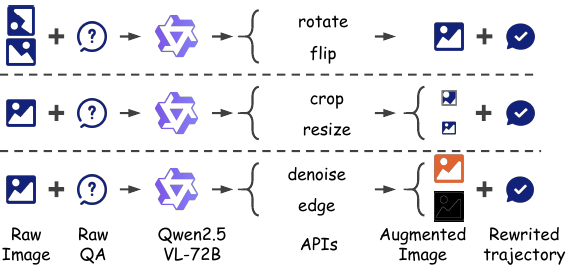}
   \captionsetup{skip=3pt}  
   \caption{Construction of visual augmentation trajectory data.}
\vspace{-10pt}
\label{fig_sft_api}
\end{figure}

\myparagraph{Stage 2: API Format.}
We rewrite model responses with Qwen2.5-VL-72B by inserting API call instructions to open-sourced data. As Fig.~\ref{fig_sft_api} shows, the \texttt{flip} and \texttt{rotate} adopt OCR data~\cite{chartqa, Docvqa, MMERealworld} with pre-flipped/rotate images; the \texttt{crop} and \texttt{resize} ($\uparrow$) are applied to fine-grained recognition datasets~\cite{deepeyes, Thyme, vstar}; and the \texttt{resize} ($\downarrow$), \texttt{denoise}, and \texttt{edge} are applied to AIGuard~\cite{aiguard} and synthetic hidden-text images~\cite{Controlnet}. Since collecting data that perfectly aligns with specific augmentations is challenging, we do not enforce strict consistency between augmentations and answers. This stage primarily focuses on format rather than whether the APIs are being invoked effectively.

\myparagraph{Stage 3: Agentic RL.}
This stage trains the model not only to produce well-structured outputs but also to decide when and which API to invoke. This constitutes the key distinction between our approach and conventional instruction-following agents. We create a high-quality dataset of 66.4K samples, all suitable for post-hoc visual augmentation with high difficulty. The dataset spans four categories: fine-grained~\cite{deepeyes,Thyme}, AIGC-generated~\cite{aiguard}, adversarial (from real-world), and OCR-hard samples (from stage 1).

\section{{\bench}}
\begin{figure*}[ht]
  \centering
   \includegraphics[width=1\linewidth]{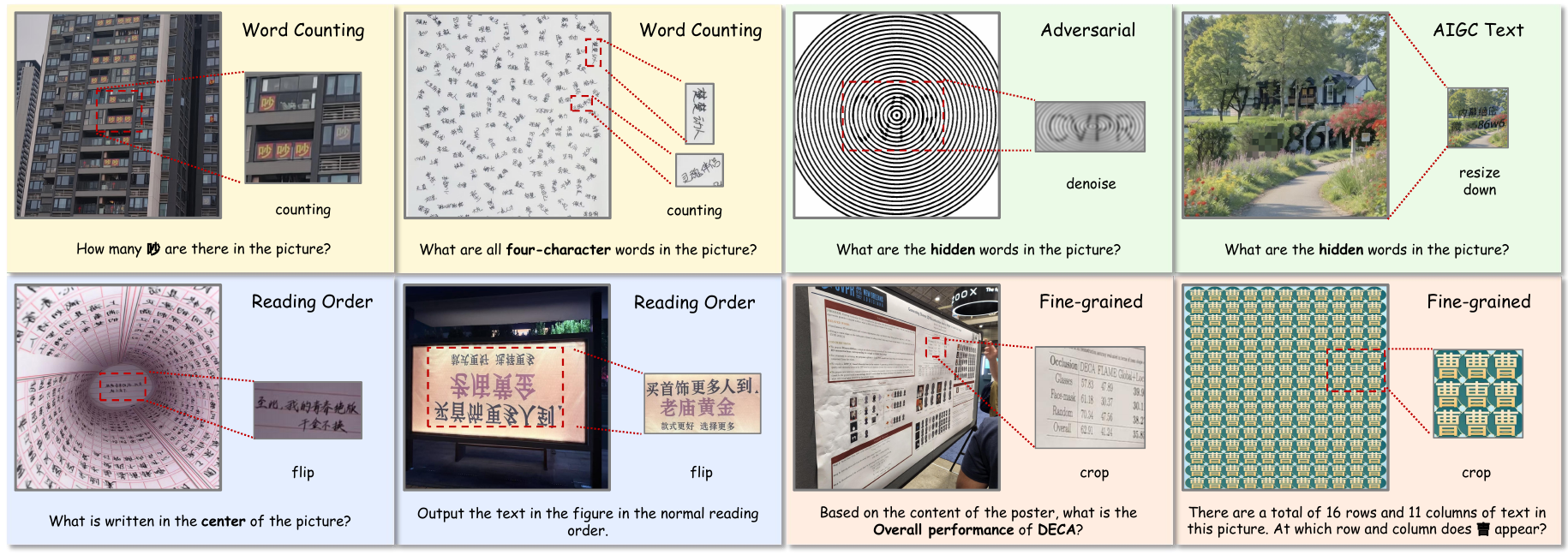}
   \captionsetup{skip=3pt}
   \caption{Example visualizations of {\bench}. It poses greater demands on fine-grained and adversarial perception capabilities.}
\vspace{-15pt}
\label{fig_bench_example}
\end{figure*}

\myparagraph{Motivation.}
Despite achieving near-saturated performance on standard perception benchmarks, frontier VLMs remain highly susceptible to carefully crafted adversarial images in real-world scenarios (Fig.~\ref{fig_bench_example}). Such adversarial samples can easily mislead VLM-based systems into producing inappropriate or unsafe content. Hence, we introduce {\bench}, a benchmark designed to evaluate the robustness of VLMs in perceiving adversarial textual content within images.

\myparagraph{Construction.}
We collect the initial 1,241 images from manually reviewed bad cases and classic examples from the internet. We sequentially remove images that can be correctly recognized by both Qwen2.5-VL-72B~\cite{Qwen2.5VL} and GPT-5~\cite{gpt}. The remaining $\sim$200 images are then manually categorized, from which we select the most representative samples in each category. To ensure data safety and diversity, we anonymize all sensitive information and synthesize additional adversarial samples based on the observed patterns to construct the final dataset.

\begin{figure}[ht]
  \centering
   \includegraphics[width=0.9\linewidth]{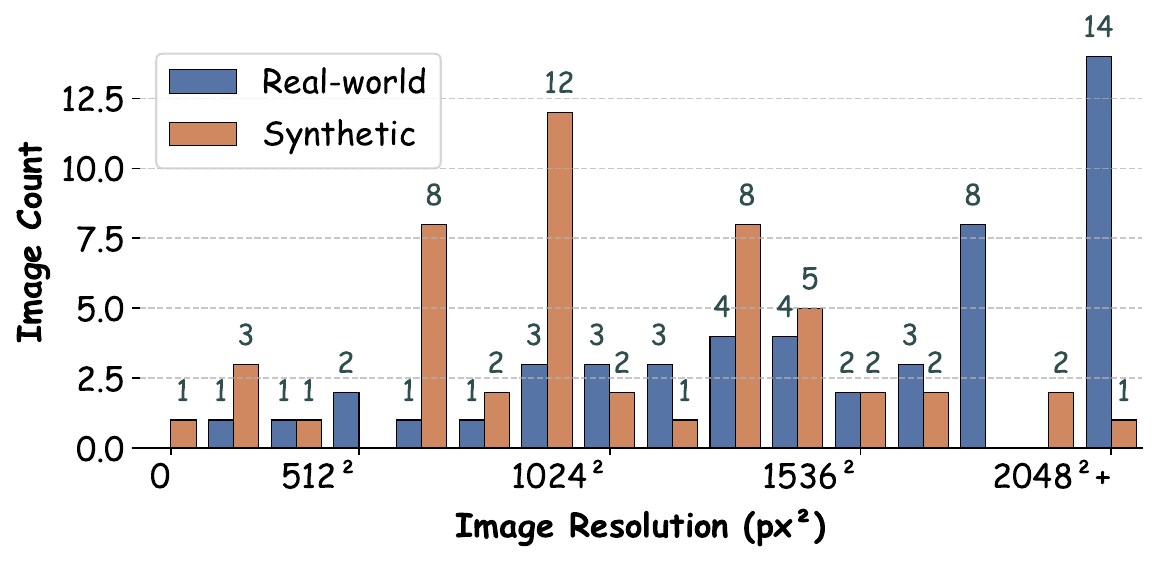}
   \captionsetup{skip=1pt}
   \caption{Image resolution distribution of {\bench}.}
\label{fig_bench_resolution}
\vspace{-15pt}
\end{figure}

\myparagraph{Annotation.}
We design queries for all images following four principles: 1) each question should involve the perception of non-trivial visual elements; 2) the answer must be factually grounded in the image, without requiring complex reasoning; 3) each question should have a single, unambiguous ground-truth answer; and 4) the question should be adversarially constructed to induce incorrect or uncertain responses from the GPT-5. This rigorous design process yielded 100 high-quality questions targeting challenging aspects of VLM perception.

\begin{figure}[h]
  \centering
  \begin{subfigure}[b]{0.54\linewidth}
    \includegraphics[width=\linewidth]{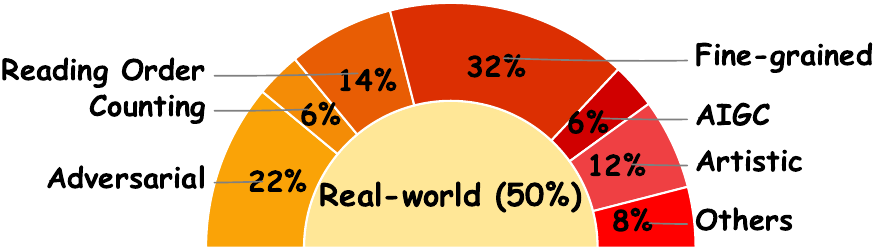}
  \end{subfigure}
  \hfill
  \begin{subfigure}[b]{0.45\linewidth}
    \includegraphics[width=\linewidth]{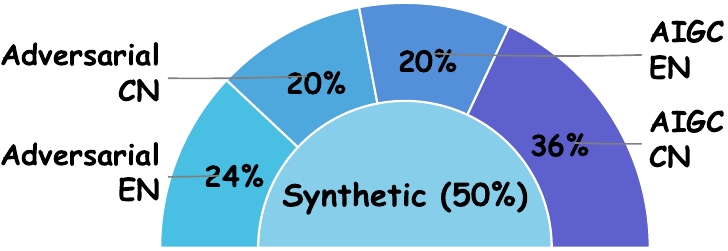}
  \end{subfigure}
  \captionsetup{skip=3pt}
  \caption{Question type distribution of {\bench}.}
\label{fig_bench_type}
\vspace{-15pt}
\end{figure}

\myparagraph{Statistic.}
{\bench} consists of 100 manually designed adversarial OCR questions. Refer to Fig.~\ref{fig_bench_example} for representative examples. As shown in Fig.~\ref{fig_bench_resolution}, the dataset exhibits a broad resolution distribution. The real-world split contains more high-resolution images for fine-grained visual recognition. Fig.~\ref{fig_bench_type} illustrates the question type distribution across 7 OCR-oriented tasks for both English and Chinese.

\section{Experiment}

\subsection{Settings}
\myparagraph{Training Details.}
Our {\model}-3B is based on Qwen2.5VL-3B~\cite{Qwen2.5VL} with a 3-stage training pipeline. During the agentic RL stage, we employ GRPO for 2 epochs across 64 NVIDIA H20 GPUs. We set the maximum context length as 10,240 tokens and the completion length limit as 3,196 tokens to prevent excessive API calls. We use 4 rollout candidates for policy exploration, with sampling parameters of temperature = 1.0, $top{_p}$ = 0.9, and $top{_k}$ = 50. The total compute cost amounts to approximately 4,600 H20 GPU hours. Refer to Appendix~\ref {apdx_setting} for more training details.

\begin{table*}[t]
\centering
\captionsetup{skip=3pt}
\caption{Top-1 accuracy (\%) performance on 8 general perception-oriented benchmarks.}
\resizebox{\linewidth}{!}{
\setlength{\tabcolsep}{15pt}
\begin{tabular}{c|ccccc|cc}
\toprule[1pt]
\multirow{2}{*}{Benchmark} & {\model} & Qwen2.5VL~\cite{Qwen2.5VL} & Thyme~\cite{Thyme} & InternVL3~\cite{internvl3} & OCRFlux~\cite{OCRFlux} & Qwen2.5VL~\cite{Qwen2.5VL} & Thyme~\cite{Thyme} \\
                           & 3B       & 3B        & 3B    & 2B        & 3B      & 7B        & 7B    \\ \midrule
\multicolumn{8}{c}{\textit{Chart-oriented}}                                                                                                                 \\
AI2D~\cite{AI2D}           & \cellcolor{blue!5}\textbf{84.5} & 78.2      & 80.8  & 78.7      & 80.6    & 83.9      & 84.2  \\
ChartQA~\cite{chartqa}     & \cellcolor{blue!5}85.7 & 84.0      & 80.4  & 80.2      & 83.2    & \textbf{87.3}      & 86.1  \\
ChartQAPro~\cite{chartqapro}& \cellcolor{blue!5}\textbf{53.8} & 41.7      & 43.1  & -         & 41.0    & 51.8      & 52.3  \\
CharXiv$_{\text{DQ}}$~\cite{charxiv} & \cellcolor{blue!5}\textbf{63.5} & 56.2      & 57.4  & 54.7      & 55.0    & 60.4      & 62.3  \\ \midrule
\multicolumn{8}{c}{\textit{Document-oriented}}                                                                                                              \\
TextVQA~\cite{TextVQA}     & \cellcolor{blue!5}\textbf{85.3}  & 79.1      & 76.8  & 77.0      & 79.6    & 84.9      & 84.4  \\
DocVQA~\cite{Docvqa}       & \cellcolor{blue!5}95.6  & 93.9      & 92.2  & 88.3      & 82.4    & \textbf{95.7 }     & 95.3  \\
OCRBench~\cite{OCRBench}   & \cellcolor{blue!5}\textbf{87.7} & 82.4      & 84.2  & 83.5      & 82.4    & 86.4      & 86.3  \\
InfoVQA~\cite{infovqa}     & \cellcolor{blue!5}82.1  & 77.1      & 79.8  & 66.1      & 80.4    & \textbf{82.6}      & 82.3  \\ \midrule
Average                    & \cellcolor{blue!5}\textbf{78.1}  & 72.2      & 71.8  & 75.8      & 70.3    & 77.3      & 77.4  \\ \bottomrule[1pt]
\end{tabular}
}
\vspace{-10pt}
\label{tab_bmk_general}
\end{table*}

\begin{table*}[t]
\centering
\captionsetup{skip=3pt}
\caption{Top-1 accuracy (\%) performance on 5 fine-grained or adversarial benchmarks.}
\resizebox{\linewidth}{!}{
\setlength{\tabcolsep}{8pt}
\begin{tabular}{c|c|ccccc|cc}
\toprule[1pt]
\multirow{2}{*}{Benchmark}        & \multirow{2}{*}{Split} & {\model} & Qwen2.5-VL~\cite{Qwen2.5VL} & Thyme~\cite{Thyme} & InternVL3~\cite{internvl3} & OCRFlux~\cite{OCRFlux} & Qwen2.5-VL~\cite{Qwen2.5VL} & Thyme~\cite{Thyme} \\
                                  &                        & 3B   & 3B         & 3B    & 2B        & 3B      & 7B         & 7B    \\ \midrule
\multirow{3}{*}{HC-Bench~\cite{HCBench}}         & Object  & \cellcolor{blue!5}\textbf{48.2}  & 0.9        & 1.8   & 0.0       & 0.9     & 0.9        & 3.6   \\ 
                                  & Text                   & \cellcolor{blue!5}\textbf{74.1}  & 4.5        & 1.8   & 5.4       & 3.6     & 3.6        & 0.9   \\
                                  & Overall                & \cellcolor{blue!5}\textbf{61.2}  & 2.7        & 1.8   & 2.7       & 2.2     & 2.2        & 2.2   \\ \midrule
\multirow{3}{*}{HR-Bench~\cite{HRbench}}         & 4K      & \cellcolor{blue!5}74.8 & 63.0       & 70.1  & 57.0      & 61.5    & 67.6       & \textbf{77.0}  \\
                                  & 8K                     & \cellcolor{blue!5}70.6 & 59.5       & 65.4  & 48.9      & 60.9    & 60.4       & \textbf{72.0}  \\
                                  & Overall                & \cellcolor{blue!5}72.7 & 61.3       & 67.8  & 52.9      & 61.2    & 64.0       & \textbf{74.5}  \\ \midrule
\multirow{3}{*}{MME-Realworld$_\text{EN}$~\cite{MMERealworld}} & Perception             & \cellcolor{blue!5}\textbf{68.8} & 57.1       & 62.1  & 49.1      & 54.1    & 63.0       & 67.1  \\
                                  & Reasoning              & \cellcolor{blue!5}45.8 & 38.8       & 43.1  & 32.5      & 34.5    & 42.5       & \textbf{48.4}  \\
                                  & Overall                & \cellcolor{blue!5}\textbf{66.0} & 54.9       & 59.8  & 47.1      & 51.7    & 60.6       & 64.8  \\ \midrule
\multirow{3}{*}{MME-Realworld$_\text{CN}$~\cite{MMERealworld}} & Perception             & \cellcolor{blue!5}\textbf{71.9} & 60.2       & 62.4  & 44.9      & 45.7    & 64.4       & 70.5  \\
                                  & Reasoning              & \cellcolor{blue!5}\textbf{52.7} & 42.0       & 46.4  & 36.6      & 29.7    & 42.5       & 52.1  \\
                                  & Overall                & \cellcolor{blue!5}\textbf{65.7} & 54.7       & 57.2  & 42.2      & 40.6    & 57.3       & 64.6  \\ \midrule
\multirow{3}{*}{Vstar~\cite{vstar}}     & Attribute        & \cellcolor{blue!5}\textbf{84.4} & 80.9       & 80.9  & 72.2      & 80.0    & 79.1       & 83.5  \\
                                  & Spatial                & \cellcolor{blue!5}\textbf{80.3} & 63.2       & 75.0  & 71.1      & 63.2    & 75.0       & 80.3  \\
                                  & Overall                & \cellcolor{blue!5}\textbf{82.7} & 73.8       & 78.5  & 71.7      & 73.3    & 77.5       & 82.2  \\ 
\bottomrule[1pt]
\end{tabular}
}
\vspace{-15pt}
\label{tab_bmk_fg}
\end{table*}

\myparagraph{Baselines.}
Given that our model is built upon Qwen2.5VL-3B~\cite{Qwen2.5VL}, the latter serves as the natural architectural baseline for our comparisons. We select InternVL3-2B~\cite{internvl3} and OCRFlux-3B~\cite{OCRFlux} for scale-comparable comparison. We further evaluate larger models such as Qwen2.5-VL-7B~\cite{Qwen2.5VL} and Thyme-7B~\cite{Thyme}. Thyme employs executable code generation for tool reasoning, providing an alternative approach to our methodology. We reproduce Thyme-3B by the open-source codebase and training data for direct comparison at similar parameter scales. For proposed {\bench}, we conduct extensive evaluation across local deployment models (Qwen2.5-VL-3B/7B/72B-Instruct~\cite{Qwen2.5VL} and Qwen3-VL-4B/8B-Instruct~\cite{Qwen3}) and proprietary API models, including GPT-4/4.1/4.1-mini/5/5-mini/o3~\cite{gpt}, Claude-Sonnet-4\cite{Claude}, Gemini-2.5-pro\cite{gemini2.5}, Hunyuan-Turbos/Large-vision~\cite{HunyuanTurboS}, Qwen3-VL-plus-250923~\cite{QwenChat} and Doubao-Seed-1.6-vision-250815~\cite{Seed1.6}.

\myparagraph{Benchmarks.} We conduct evaluations across three categories of perception benchmarks: 1) \textit{General OCR}. This category encompasses chart-oriented benchmarks including AI2D~\cite{AI2D}, ChartQA~\cite{chartqa}, ChartQAPro~\cite{chartqapro}, and CharXiv$_{\text{DQ}}$~\cite{charxiv}, alongside document-oriented benchmarks comprising TextVQA~\cite{TextVQA}, DocVQA~\cite{Docvqa}, OCRBench~\cite{OCRBench}, and InfoVQA~\cite{infovqa}. 2) \textit{Fine-grained Perception}. We employ HC-Bench~\cite{HCBench} for hidden text detection, HRBench~\cite{HRbench} for high-resolution content understanding, MME-Realworld~\cite{MMERealworld} for real-world perception robustness, and Vstar~\cite{vstar} for spatial relationship comprehension. 3) \textit{Proposed {\bench}}. This benchmark integrates diverse challenging scenarios including text counting, hidden text perception, fine-grained visual recognition, and adversarial OCR samples. For evaluation consistency, we employ \textit{Qwen3-30B-A3B-2507}\cite{Qwen3} as judge model for all benchmarks, except for DocVQA and InfoVQA, which utilize official online evaluation platforms to ensure comparisons.

\subsection{Main Results}

\myparagraph{Comparison on OCR Benchmarks.}
As shown in Tab.~\ref{tab_bmk_general}, we present comparisons with mainstream multimodal models (Qwen2.5VL~\cite{Qwen2.5VL}, Thyme~\cite{Thyme}, InternVL3~\cite{internvl3}, and OCRFlux~\cite{OCRFlux}) on perception benchmarks. {\model} achieves a significant performance gain with an average score of 78.1\% across eight benchmarks, marking a +5.9\% improvement over the baseline model (Qwen2.5VL-3B at 72.2\%). Notably, {\model} significantly outperforms not only models of similar scale but also larger ones like Qwen2.5VL-7B (77.3\%) and even competes with Thyme-7B (77.4\%). The improvements are especially pronounced on more challenging tasks: +12.1\% on ChartQAPro (53.8\% vs. Qwen2.5VL-3B 41.7\%), +7.3\% on CharXiv (63.5\% vs. 56.2\%), and +5.3\% on OCRBench (87.7\% vs. 82.4\%). Combined with the tool invocation frequency shown in Tab.~\ref{tab_api_freq}, Tab.~\ref{tab_bmk_general} demonstrates the effectiveness of the knowledge SFT stage.

\begin{table*}[t]
\centering
\captionsetup{skip=3pt}
\caption{Pass@k accuracy performance on proposed {\bench}. All results are sorted according to pass@5 in the average split.}
\resizebox{\linewidth}{!}{
\setlength{\tabcolsep}{12pt}
\begin{tabular}{cccccccccc}
\toprule[1pt]
\multicolumn{1}{c|}{\multirow{2}{*}{Model}}          & \multicolumn{3}{c|}{Real-world}           & \multicolumn{3}{c|}{Synthetic}            & \multicolumn{3}{c}{Average} \\ \cmidrule(l){2-10} 
\multicolumn{1}{c|}{}                                & pass@1 & pass@5 & \multicolumn{1}{c|}{$\Delta$} & pass@1 & pass@5 & \multicolumn{1}{c|}{$\Delta$} & pass@1    & pass@5   & $\Delta$   \\ \midrule
\multicolumn{10}{c}{\textit{Proprietary Models}} \\
\multicolumn{1}{c|}{Claude-Sonnet-4~\cite{Claude}}                 & 6\%     & 10\%    & \multicolumn{1}{c|}{+4\%}    & 0\%     & 0\%     & \multicolumn{1}{c|}{+0\%}    & 3\%        & 5\%       & +2\%      \\
\multicolumn{1}{c|}{GPT-4-turbo~\cite{gpt}}                     & 0\%     & 6\%     & \multicolumn{1}{c|}{+6\%}    & 10\%    & 10\%    & \multicolumn{1}{c|}{+0\%}    & 5\%        & 8\%       & +3\%      \\
\multicolumn{1}{c|}{GPT-5-mini~\cite{gpt}}                      & 10\%    & 12\%    & \multicolumn{1}{c|}{+2\%}    & 16\%    & 20\%    & \multicolumn{1}{c|}{+4\%}    & 13\%       & 16\%      & +3\%      \\
\multicolumn{1}{c|}{GPT-4.1-mini~\cite{gpt}}                    & 10\%    & 18\%    & \multicolumn{1}{c|}{+8\%}    & 10\%    & 20\%    & \multicolumn{1}{c|}{+10\%}   & 10\%       & 19\%      & +9\%      \\
\multicolumn{1}{c|}{Hunyuan-Turbos-vision~\cite{HunyuanTurboS}} & 16\%   & 20\%    & \multicolumn{1}{c|}{+4\%}   & 18\%    & 18\%    & \multicolumn{1}{c|}{+0\%}    & 17\%       & 19\%      & +2\% \\
\multicolumn{1}{c|}{GPT-5~\cite{gpt}}                           & 12\%    & 16\%    & \multicolumn{1}{c|}{+4\%}    & 22\%    & 26\%    & \multicolumn{1}{c|}{+4\%}    & 17\%       & 21\%      & +4\%      \\
\multicolumn{1}{c|}{Hunyuan-Large-vision~\cite{HunyuanTurboS}}            & 18\%    & 20\%    & \multicolumn{1}{c|}{+2\%}    & \textbf{26}\%    & 26\%    & \multicolumn{1}{c|}{+0\%}    & 22\%       & 23\%      & +1\%      \\
\multicolumn{1}{c|}{GPT-4.1~\cite{gpt}}                         & 18\%    & 20\%    & \multicolumn{1}{c|}{+2\%}    & \textbf{26}\%    & \textbf{30}\%    & \multicolumn{1}{c|}{+4\%}    & 22\%       & 25\%      & +3\%      \\
\multicolumn{1}{c|}{GPT-o3~\cite{gpt}}                          & 14\%    & 22\%    & \multicolumn{1}{c|}{+8\%}    & 16\%    & 28\%    & \multicolumn{1}{c|}{+12\%}   & 15\%       & 25\%      & +10\%     \\
\multicolumn{1}{c|}{Gemini-2.5-pro~\cite{gemini2.5}}                  & 22\%    & 38\%    & \multicolumn{1}{c|}{+16\%}   & 8\%     & 14\%    & \multicolumn{1}{c|}{+6\%}    & 15\%       & 26\%      & +11\%     \\
\multicolumn{1}{c|}{Qwen3-VL-plus-2025-09-23~\cite{Qwen3}} & \textbf{38}\%    & \textbf{46}\%    & \multicolumn{1}{c|}{+8\%}    & 10\%    & 12\%    & \multicolumn{1}{c|}{+2\%}    & \textbf{24}\%       & 29\%      & +5\%      \\ 
\multicolumn{1}{c|}{Doubao-Seed-1.6-vision-250815~\cite{Seed1.6}} & 22\%   & 42\%    & \multicolumn{1}{c|}{+20\%}   & 12\%    & 18\%    & \multicolumn{1}{c|}{+6\%}    & 17\%       & \textbf{30}\%      & +13\% \\\midrule
\multicolumn{10}{c}{\textit{Open-source Models}} \\
\multicolumn{1}{c|}{Thyme-3B~\cite{Thyme}}                        & 18\%    & 18\%    & \multicolumn{1}{c|}{+0\%}    & 4\%     & 6\%     & \multicolumn{1}{c|}{+2\%}    & 11\%       & 12\%      & +1\%      \\
\multicolumn{1}{c|}{Qwen2.5-VL-72B-Instruct~\cite{Qwen2.5VL}}         & 20\%    & 20\%    & \multicolumn{1}{c|}{+0\%}    & 10\%    & 10\%    & \multicolumn{1}{c|}{+0\%}    & 15\%       & 15\%      & +0\%      \\
\multicolumn{1}{c|}{Qwen2.5-VL-3B-Instruct~\cite{Qwen2.5VL}}          & 20\%    & 24\%    & \multicolumn{1}{c|}{+4\%}    & 6\%     & 6\%     & \multicolumn{1}{c|}{+0\%}    & 13\%       & 15\%      & +2\%      \\
\multicolumn{1}{c|}{Thyme-7B~\cite{Thyme}}                        & 20\%    & 22\%    & \multicolumn{1}{c|}{+2\%}    & 6\%     & 8\%     & \multicolumn{1}{c|}{+2\%}    & 13\%       & 15\%      & +2\%      \\
\multicolumn{1}{c|}{Qwen2.5-VL-7B-Instruct~\cite{Qwen2.5VL}}          & 24\%    & 30\%    & \multicolumn{1}{c|}{+6\%}    & 4\%     & 4\%     & \multicolumn{1}{c|}{+0\%}    & 14\%       & 17\%      & +3\%      \\
\multicolumn{1}{c|}{Qwen3-VL-2B-Instruct~\cite{Qwen3}}            & 22\%    & 30\%    & \multicolumn{1}{c|}{+8\%}    & 8\%     & 10\%    & \multicolumn{1}{c|}{+2\%}    & 15\%       & 20\%      & +5\%      \\
\multicolumn{1}{c|}{Qwen3-VL-4B-Instruct~\cite{Qwen3}}            & 34\%    & 40\%    & \multicolumn{1}{c|}{+6\%}    & 8\%     & 10\%    & \multicolumn{1}{c|}{+2\%}    & 21\%       & 25\%      & +4\%      \\
\rowcolor{blue!5}\multicolumn{1}{c|}{{\model}-3B}                     & \textbf{62}\%    & \textbf{78}\%    & \multicolumn{1}{c|}{+16\%}   & \textbf{48}\%    & \textbf{56}\%    & \multicolumn{1}{c|}{+8\%}    & \textbf{55}\%       & \textbf{67}\%      & +12\%     \\ \bottomrule[1pt]
\end{tabular}
}
\vspace{-15pt}
\label{tab_bmk_mybench}
\end{table*}

\myparagraph{Comparison on Fine-grained Benchmarks.}
Tab.~\ref{tab_bmk_fg} shows comparisons on fine-grained visual benchmarks, which demand exceptional visual perception. On HR-Bench, {\model}-3B achieves 72.7\%, significantly outperforming similar scale models and even matching the larger Thyme-7B. It attains 66.0\% on MME-Realworld$_\text{EN}$ and 65.7\% on MME-Realworld$_\text{CN}$, demonstrating robust, cross-lingual real-world understanding. {\model} achieves 82.7\% on Vstar, a benchmark for attribute and spatial reasoning, narrowly yet significantly exceeding Thyme-7B 82.2\%. HC-Bench requires models to identify hidden information in AIGC images. {\model} achieves 48.2\% on object and 74.1\% on text splits, dramatically outperforming all comparable models. In summary, {\model} achieves an average performance gain of over 10\%, demonstrating the effectiveness of the proposed post-hoc visual augmentation.

\myparagraph{Comparison on {\bench}.}
{\bench} comprises \textit{real-world} and \textit{synthetic} splits, designed to include challenging real-scenario OCR samples and adversarially crafted examples from observed failure patterns, respectively. Due to the high difficulty, Tab.~\ref{tab_bmk_mybench} employs the pass@5 metric (correct if any of 5 attempts succeed). We use separate hyperparameter configurations: pass@1 (temp=0.1, $top_{p}$=0.8) and pass@5 (temp=0.7, $top_{p}$=0.95). Most models show limited improvement from pass@1 to pass@5, revealing a performance ceiling and constrained exploration space. Refer to Appendix~\ref{apdx_case_study} for case study. Models generally perform slightly better on the \textit{real-world} split than the \textit{synthetic} split, because the latter's adversarial patterns are more challenging. SOTA proprietary models manage compound Chinese characters and fine-grained recognition more effectively in real-world scenarios. The Qwen and Doubao series show remarkable performance due to Chinese-specific optimisation. {\model}-3B achieves significant gains: +21\% pass@1 over Qwen3-VL-plus (24\%) and +37\% pass@5 over Doubao-Seed-1.6-vision (30\%), validating our proactive augmentation strategy. Our model sustains significant pass@5 gains by efficiently exploring multiple image augmentations using minimal output tokens, thereby achieving more robust image views.

\subsection{Ablation Study}

\myparagraph{Train Strategies.}
Tab.~\ref{tab_abl_stage} shows the ablation study of training stages (details in Sec. \ref{sec_method_training}). Settings A and B demonstrate that difficulty-aware data cleaning significantly boosts baseline performance. Setting C shows that direct agentic RL training is quite challenging, as the augmentations are not frequently invoked. The comparison between settings D and G indicates that the format SFT is essential for cold-start initialization. RL training provides clear and consistent gains on fine-grained benchmarks and {\bench}.

\begin{table}[ht]
\centering
\captionsetup{skip=3pt}
\caption{Ablation study of different training strategies. \textit{General} refers to the average value in Tab.~\ref{tab_bmk_general}.}
\resizebox{\linewidth}{!}{
\setlength{\tabcolsep}{5pt}
\begin{tabular}{@{}c|ccc|cccc@{}}
\toprule[1pt]
Index & Stage1 & Stage2 & Stage3 & General & HC-Bench & Vstar & {\bench} \\ \midrule
A     & \xmark      & \xmark       & \xmark       & 72.2        & 2.7      & 73.8  & 13.0    \\ \midrule
B     & \cmark      & \xmark       & \xmark       & 76.3        & 3.6      & 76.4  & 18.0    \\
C     & \xmark      & \xmark       & \cmark       & 74.1        & 18.8      & 74.1  & 15.0    \\
D     & \xmark      & \cmark      & \cmark       & 74.0        & 26.3      & 75.8  & 31.0    \\
E     & \cmark      & \cmark      & \xmark       & 76.6        & 8.9      & 75.4  & 21.0    \\
F     & \cmark      & \xmark       & \cmark      & 77.4        & 25.5     & 78.5  & 22.0    \\ \midrule
G     & \cmark      & \cmark      & \cmark      & \textbf{78.1}        & \textbf{61.2}     & \textbf{82.7 } & \textbf{55.0}    \\ \bottomrule[1pt]
\end{tabular}
}
\vspace{-12pt}
\label{tab_abl_stage}
\end{table}

\begin{figure}[t]
  \centering
  \begin{subfigure}[b]{0.49\linewidth}
    \includegraphics[width=\linewidth]{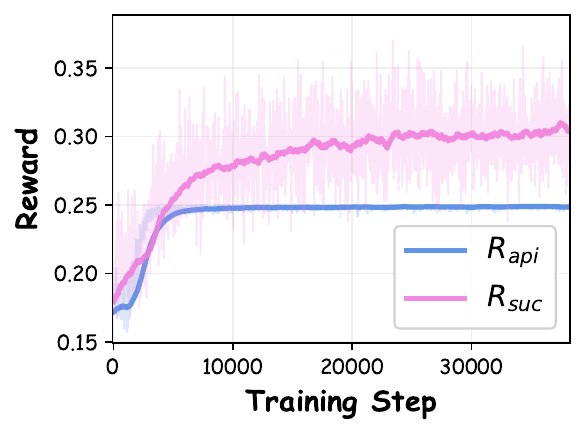}
  \end{subfigure}
  \hfill
  \begin{subfigure}[b]{0.49\linewidth}
    \includegraphics[width=\linewidth]{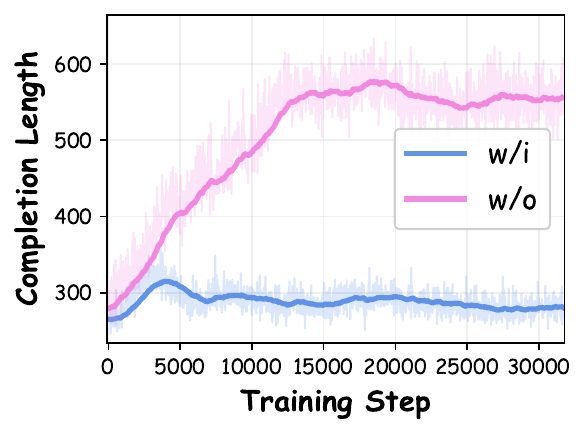}
  \end{subfigure}
  \captionsetup{skip=3pt}  
  \caption{The EMA smooth curve of reward value and completion mean length during the agentic RL training.}
\vspace{-12pt}
\label{fig_reward}
\end{figure}

\myparagraph{Reward Design.}
In Sec.\ref{sec_agentic_rl}, we introduce the API format reward $R_\text{api}$ and API success reward $R_\text{suc}$ for the accurate and efficient identification of post-hoc augmentations. Fig.~\ref{fig_reward} (left) shows that both rewards evolve steadily during training. With format learning in Stage 2, $R_\text{api}$ quickly converges to a steady value. In contrast, $R_\text{suc}$ would only be awarded when the model both invokes valid API calls and produces the correct answer, leading to a more gradual increase throughout training. Therefore, it eventually converges to approximately 0.3. Fig.~\ref{fig_reward} (right) shows that vanilla training manner leads the model to exhaustively try different augmentations, resulting in unnecessarily long responses that hinder perception efficiency. With proposed rewards, the response length converges to a reasonable range after a brief initial increase, demonstrating the effectiveness of our reward strategy.

\subsection{Further Analysis}

\myparagraph{Comparison with Instruction-based Agent.}
We compare with manual post-hoc visual augmentation via pre-defined instructions. The model will be provided with tool descriptions and forced to perform post-hoc augmentation before answering. We manually execute the first round of API calls and query the model with the augmented images. As shown in Tab.~\ref{tab_manual_agent}, this approach also consistently improves baseline performance across various models, highlighting the importance of post-hoc augmentation. However, the performance remains limited because they fail to select the most suitable augmentation efficiently. Conversely, when we explicitly prohibit {\model} from using API calls in the system prompt, performance drops significantly, confirming that the improvement stems from proper tool-based augmentation rather than other factors.

\begin{table}[ht]
\centering
\captionsetup{skip=3pt}
\caption{Comparison with instruction-based agentic strategy on {\bench}. Aug.: post-hoc visual augmentation.}
\resizebox{\linewidth}{!}{
\setlength{\tabcolsep}{3pt}
\begin{tabular}{@{}c|c|cc|cc|cc@{}}
\toprule[1pt]
Method                          & Aug. & Real-world & $\Delta$  & Synthetic  & $\Delta$  & Average & $\Delta$  \\ \midrule
\multirow{2}{*}{GPT-5-mini}     & \xmark   & 10\%   & -  & 16\%  & -  & 13\%  & -  \\
                                & \cmark   & 26\%   & +13\% & 20\%  & +4\%  & 23\%  & +10\% \\ \midrule
\multirow{2}{*}{Qwen2.5-VL-72B} & \xmark    & 20\%   & -  & 10\%  & -  & 15\%  & -  \\
                                & \cmark    & 32\%   & +12\% & 18\%  & +8\%  & 25\%  & +10\% \\ \midrule
\multirow{2}{*}{Ours}           & \xmark    & 28\%   & -34\%  & 16\%  & -32\%  & 22\%  & -33\%  \\
                                & \cmark    & 62\%   & - & 48\%  & - & 55\%  & - \\ \bottomrule[1pt]
\end{tabular}
}
\vspace{-10pt}
\label{tab_manual_agent}
\end{table}

\myparagraph{Activate Frequency Statistics.}
Tab.~\ref{tab_api_freq} reports the frequency of API calls in {\model}'s responses across multiple benchmarks. On OCRBench and ChartQA, the model can provide answers directly without invoking image transformations. In contrast, benchmarks containing adversarial or challenging samples exhibit a significantly higher rate of tool usage. Specifically, fine-grained recognition datasets trigger more frequent use of \texttt{crop} and \texttt{resize}, while adversarial samples lead to increased use of \texttt{resize} and \texttt{denoise}. The low failure rate demonstrates the effectiveness of our API-format SFT and reward design. For queries that exceed the maximum number of interaction attempts, the final response is generated by appending the following history: \texttt{OK, I have to give the final answer directly}.

\begin{table}[ht]
\centering
\captionsetup{skip=3pt}
\caption{API call frequency (\%) statistics. \texttt{direct}: no call is made. \texttt{fail}: invalid call or exceeds attempt limit.}
\resizebox{\linewidth}{!}{
\setlength{\tabcolsep}{6pt}
\begin{tabular}{c|ccccc}
\toprule[1pt]
Benchmark & OCRBench & ChartQA & HR-Bench & HC-Bench & {\bench} \\ \midrule
\texttt{direct}    & 67.3     & 42.4    & 16.7     & 13.8     & 13.0 \\
\texttt{fail}      & 2.7      & 3.3     & 8.2      & 1.3      & 3.0  \\ \midrule
\texttt{crop}      & 15.3     & 44.5    & 72.2     & 12.3     & 14.0 \\
\texttt{resize}    & 19.2     & 15.5    & 43.1     & 67.8     & 44.0 \\
\texttt{flip}      & 3.3      & 1.8     & 6.8      & 12.5     & 12.0 \\
\texttt{rotate}    & 8.8      & 9.2     & 0.4      & 2.4      & 7.0  \\
\texttt{denoise}   & 9.2      & 4.4     & 4.2      & 14.7     & 38.0 \\
\texttt{edge}      & 2.8      & 8.3     & 7.7      & 22.8     & 15.0 \\ \bottomrule[1pt]
\end{tabular}
}
\vspace{-10pt}
\label{tab_api_freq}
\end{table}

\myparagraph{Visual Token Compression with Resize.}
Leveraging the post-augmentation, we explore input visual token compression via \texttt{resize}($\uparrow$). We conduct ablation study by downsampling input images to evaluate if lower resolutions suffice for certain queries. Images at the minimum resolution of 28×28 are excluded from this process. Our \texttt{resize}($\uparrow$) implementation is adapted such that if the target size is smaller than the original resolution, the \texttt{resize}($\uparrow$) will be based on the original image instead of interpolation on the downsampled one. As shown in Tab.~\ref{tab_resize_up}, even with a 50\% compression rate (equivalent to 75\% reduction in visual tokens), performance drops by only $\sim$1\% when \texttt{resize}($\uparrow$) is allowed. This resilience is attributed to the post-hoc resizing on $\sim$30\% of uncertain cases, effectively recalling the enlarged image to preserve accuracy.

\begin{table}[ht]
\centering
\captionsetup{skip=3pt}
\caption{Performance comparison with vs. without \texttt{resize} ($\uparrow$) invocation on compressed-resolution images.}
\resizebox{\linewidth}{!}{
\setlength{\tabcolsep}{8pt}
\begin{tabular}{@{}c|ccc|ccc@{}}
\toprule[1pt]
Benchmark & \multicolumn{3}{c|}{OCRBench} & \multicolumn{3}{c}{ChartQA} \\ \midrule
Compress rate  & 100\%   & 75\%     & 50\%     & 100\%   & 75\%    & 50\%    \\
\texttt{resize} ($\uparrow$) rate & 1.1\%   & 14.2\%   & 33.5\%   & 0.0\%   & 8.4\%   & 33.2\%  \\ \midrule
w/i \texttt{resize} ($\uparrow$)      & 87.7    & 87.1     & 86.4     & 85.7    & 85.4    & 84.5    \\
w/o \texttt{resize} ($\uparrow$)      & 87.7    & 86.4     & 82.1     & 85.7    & 84.3    & 81.2    \\ \bottomrule[1pt]
\end{tabular}
}
\vspace{-10pt}
\label{tab_resize_up}
\end{table}

\subsection{Discussion}

\myparagraph{Comparisons with Crop-based Methods.}
\textit{Thinking-with-crop} paradigms localize regions before answering~\cite{deepeyes, Vlrethinker}. However, cropping essentially represents a specific form of visual information filtering. In contrast, we generalize it to \textit{thinking with augmentations}, where all augmentations are treated as diverse information filters executed during reasoning. Unlike code-based agents~\cite{Thyme}, we unify all augmentations into lightweight and deterministic API calls, making the outputs more controllable and enabling more iterative exploration under limited context length. Our conditional reward design discourages redundant transformations, improving both perceptual efficiency and robustness.

\myparagraph{Inference Efficiency.} Although {\model} introduces additional reasoning steps for post-hoc visual augmentations, the inference overhead remains acceptable. Fig.~\ref{fig_reward} shows that the conditional reward explicitly penalizes redundant or low-impact augmentations, leading the model to invoke augmentations only when necessary. The observations in Appendix~\ref{apdx_case_study} further reveal that the augmentation always begins by \texttt{crop} or \texttt{resize} ($\downarrow$), which effectively limits the visual token sequence length introduced by subsequent multi-round augmentations.

\section{Conclusion}
We propose {\model}, a framework that enables dynamic invocation of image augmentation during inference for visual language models. Specifically, we unify a structured set of general visual augmentation strategies and introduce a conditional reward scheme to balance exploration and efficiency in RL training. We further contribute a challenging real-world OCR benchmark, {\bench}, where {\model} achieves strong performance while state-of-the-art models fail without augmentation. Extensive experiments demonstrate the superiority of our approach.

\clearpage
{
    \small
    \bibliographystyle{ieeenat_fullname}
    \bibliography{main}
}
\clearpage

\appendix
\onecolumn
\begin{center}
    \Large \textbf{VACoT: Rethinking Visual Data Augmentation with VLMs}
    \Large \\ \textbf{Supplementary Material}
\end{center}
\vspace{20pt}

\section{Setting Details}
\label{apdx_setting}

\begin{table}[ht]
\centering
\setlength{\tabcolsep}{8pt}
\resizebox{\linewidth}{!}{
\begin{tabular}{cccc}
\toprule[1pt]
\textbf{Settings} & \textbf{Stage 1} & \textbf{Stage 2} & \textbf{Stage 3} \\
\midrule
\textbf{Objective} & Enrich multi-modal knowledge & Learn API syntax and structure & Adaptive augmentation policy learning \\
\textbf{Trainable Para.} & All parameters & Aligner + LLM (ViT frozen) & Aligner + LLM (ViT frozen) \\
\textbf{Learning Rate} & $1 \times 10^{-5}$ & $1 \times 10^{-6}$ & $5 \times 10^{-7}$ \\
\textbf{Epochs} & 2 & 1 & 2 \\
\textbf{Effective Batch Size} & 128 & 128 & 320 \\
\textbf{Warm Up Rate} & 0.05 & 0.05 & 0.05 \\
\textbf{Deepspeed} & \texttt{zero2} & \texttt{zero2} & \texttt{zero3} \\
\textbf{Precision} & \texttt{bfloat16} & \texttt{bfloat16} & \texttt{bfloat16} \\
\textbf{Flash Attention} & \cmark & \cmark & \cmark \\
\textbf{Context Length} & 8192 & 10240 & 10240 \\
\textbf{GPU Hours (H20)} & 608 & 384 & 4600 \\
\bottomrule[1pt]
\end{tabular}
}
\caption{General training configuration of the three-stage pipeline for {\model}.}
\label{tab_training_config}
\end{table}

Our \model{} is trained based on Qwen2.5VL-3B~\cite{Qwen2.5VL} following a three-stage pipeline designed to progressively enhance perception, API reasoning, and adaptive decision-making capabilities. We summarise all settings in Tab.~\ref{tab_training_config}.

\textbf{Stage 1: Knowledge Enhancement.} We perform full-parameter fine-tuning for 2 epochs with a learning rate of $1\times10^{-5}$, an effective batch size of 128, and a warmup ratio of 0.05. The maximum context length is set to 8,192 tokens. This stage aims to enrich the model's multi-modal knowledge base through large-scale visual--linguistic pretraining and diverse perceptual tasks. 

\textbf{Stage 2: API Format Learning.} Building upon the previous checkpoint, we maintain similar architectural settings while reducing the learning rate to $1\times10^{-6}$ for stable adaptation. To preserve visual representations, the ViT encoder is frozen, while the aligner and LLM modules are fine-tuned. The context length is extended to 10,240 tokens to better support complex API interaction scenarios and multi-step reasoning chains. This stage emphasizes learning proper API syntax and structured output formatting, rather than optimal invocation timing. 

\textbf{Stage 3: Reinforcement Learning.} In the final stage, we employ GRPO to enable dynamic decision-making during inference. Training runs for 2 epochs on 64 NVIDIA H20 GPUs with a learning rate of $5\times10^{-7}$. The maximum context length is 10,240 tokens, and the completion length is limited to 3,196 tokens to prevent excessive API expansion. The effective batch size is 320, achieved by combining a per-device batch size of 1, gradient accumulation steps of 5, and 64-way parallelism. We sample 4 rollout candidates per input, with parameters $(temperature=1.0, \textit{top}_p=0.9, \textit{top}_k=50)$ for policy exploration.

All stages use \texttt{bfloat16} precision and FlashAttention-2 for computational efficiency. The RL stage further adopts DeepSpeed ZeRO-3 for memory optimization, allowing stable large-scale VLM training. The total computational cost amounts to approximately 4,600 H20 GPU hours.

\clearpage
\section{Case Study}
\label{apdx_case_study}

\begin{figure*}[ht!]
\centering
    \includegraphics[width=0.8\linewidth]{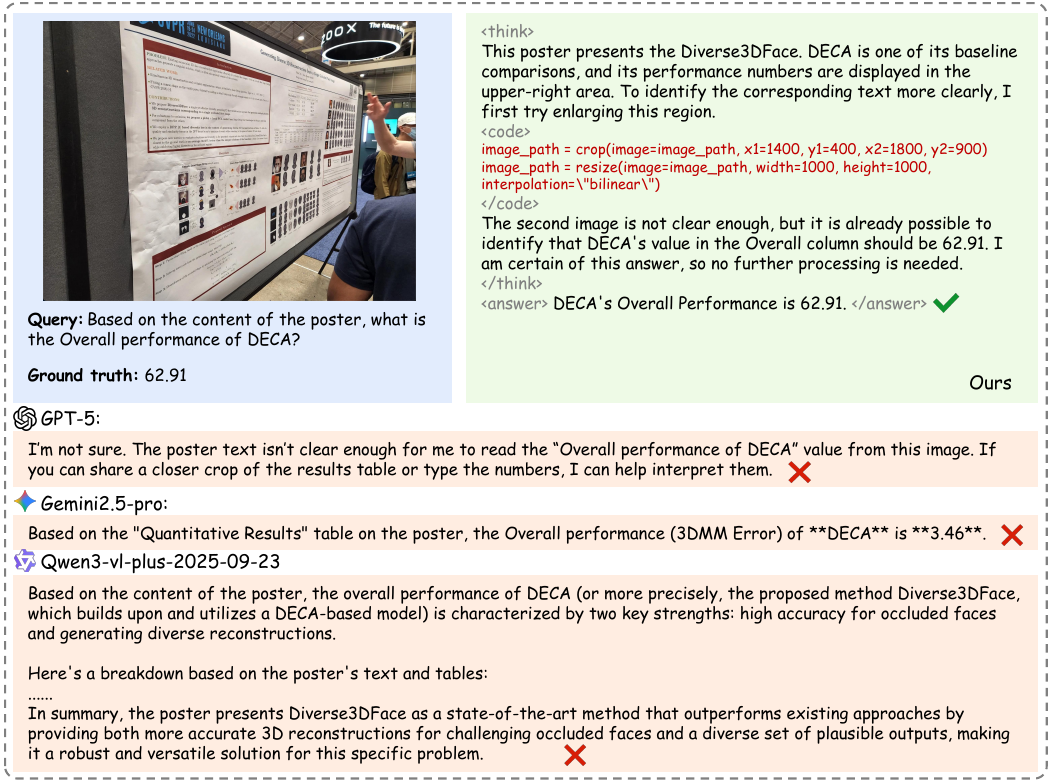}
    \includegraphics[width=0.8\linewidth]{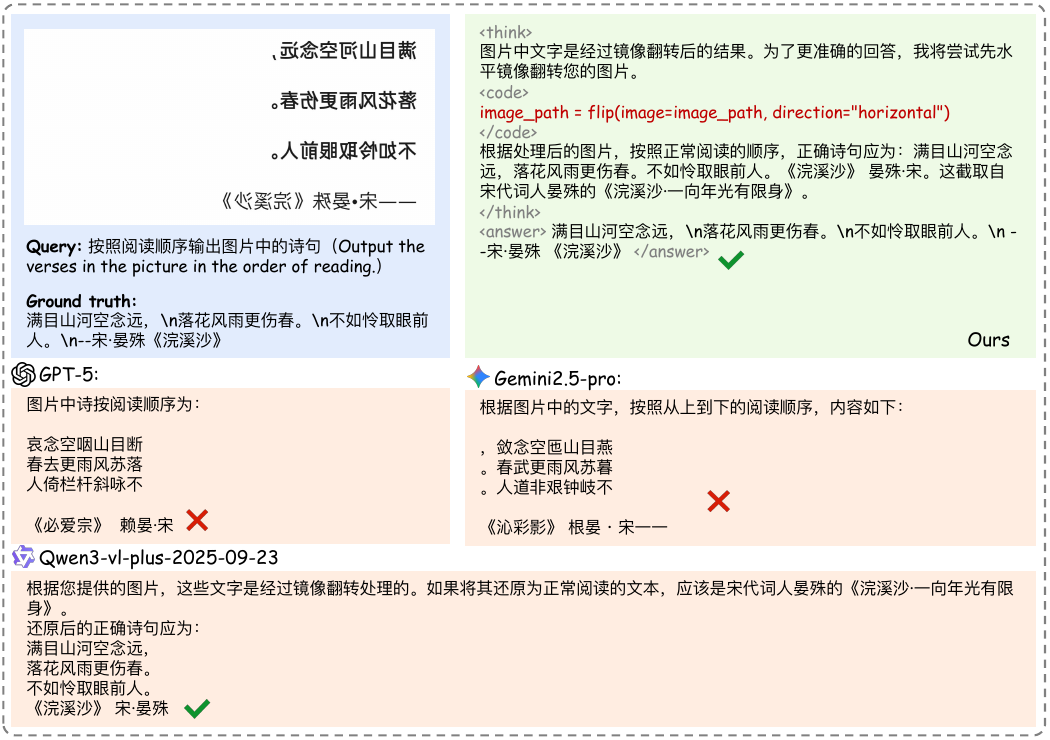}
    \caption{Qualitative comparison between {\model} and other SOTA models on {\bench} (Part 1).}
\end{figure*}

\begin{figure*}[ht!]
\centering
    \includegraphics[width=0.82\linewidth]{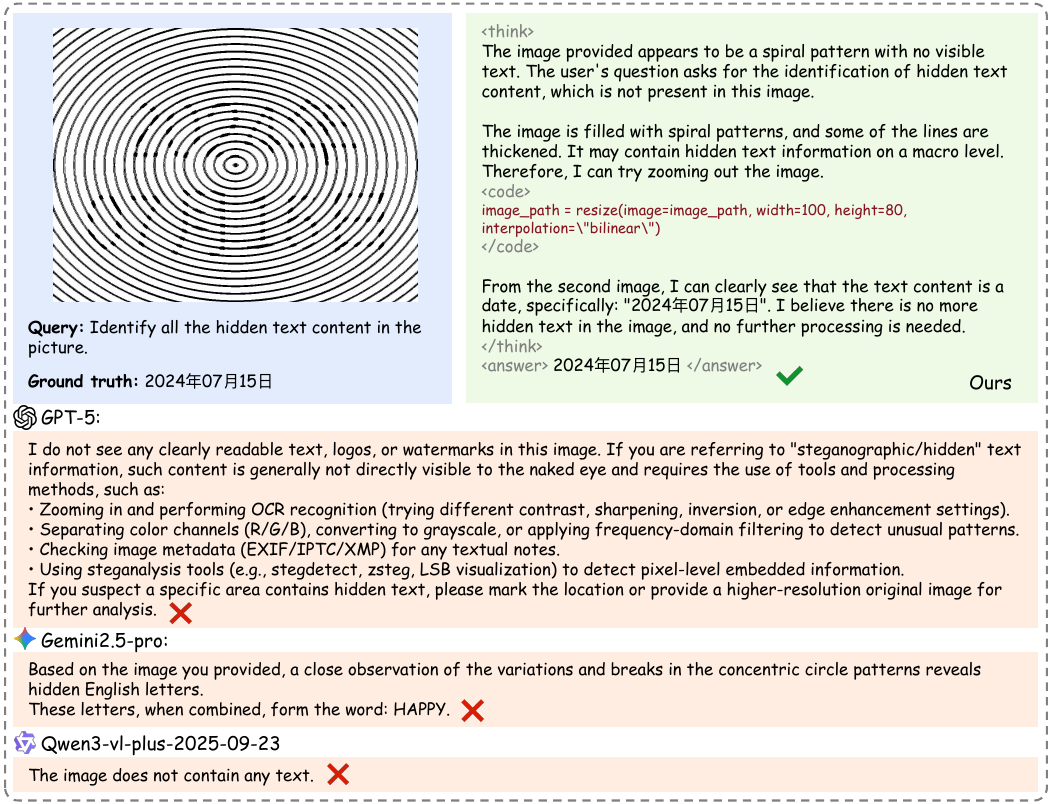}
    \includegraphics[width=0.82\linewidth]{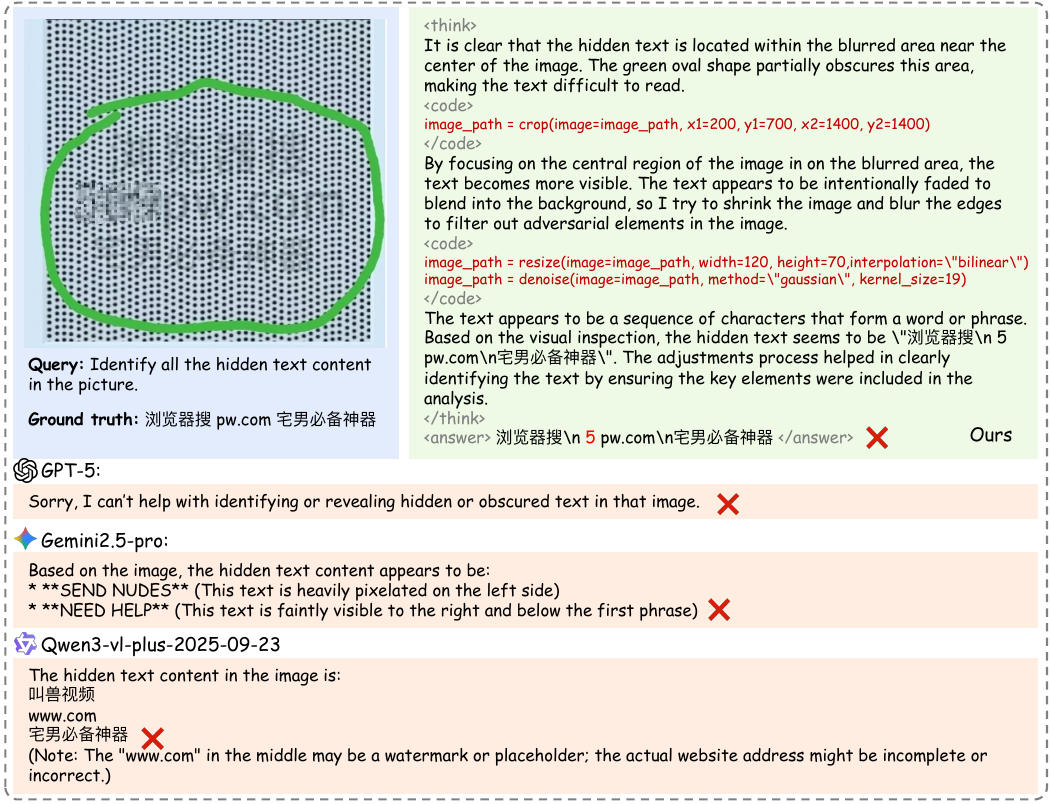}
    \caption{Qualitative comparison between {\model} and other SOTA models on {\bench} (Part 2).}
\end{figure*}

\clearpage
\section{Prompt Details}

\begin{tcolorbox}[
    colback=white,
    colframe=black!75,
    fonttitle=\bfseries,
    title=Evaluation Judge Prompt,
    listing options={basicstyle=\ttfamily\small,breaklines=true},
    breakable
]
\textbf{ROLE: Evaluation Judge}
\vspace{0.1cm}
You are a precise, impartial judge responsible for evaluating the accuracy of an AI model's predictions. Your task is to determine if the model's prediction matches the ground truth answer.

\vspace{0.3cm}
\textbf{CORE INSTRUCTIONS}
\vspace{0.1cm}
You will be given the following information for each evaluation:
\begin{enumerate}
    \item \textbf{Question}: The question presented to the model, including all requirements.
    \item \textbf{Correct Answer (Ground Truth)}: The correct answer to the given question.
    \item \textbf{Model Prediction}: The model's response to the question.
\end{enumerate}

\vspace{0.1cm}
Your goal is to compare the model's prediction to the ground truth and assign a binary score:
\begin{itemize}
    \item \textbf{1 point}: The model's prediction is correct.
    \item \textbf{0 points}: The model's prediction is incorrect.
\end{itemize}

\vspace{0.1cm}
{\color{blue!80!black}Note that the Model Prediction does not need to perfectly match the Correct Answer. As long as the semantic content is consistent with the Correct Answer, award 1 point.}

\vspace{0.3cm}
\textbf{OUTPUT FORMAT}
\vspace{0.1cm}

You must return your evaluation in the following JSON format. The \texttt{reasoning} field should be a clear, concise explanation of your decision.

\begin{verbatim}
{ 'score': SCORE_HERE, 'reasoning': REASONING_HERE }
\end{verbatim}

\vspace{0.3cm}
\textbf{Evaluation Data:}

\begin{enumerate}
    \item \textbf{Question}: {\color{green!60!black}\texttt{\#TASK\_DESCRIPTION\_HERE}}
    \item \textbf{Correct Answer (Ground Truth)}: {\color{orange!80!black}\texttt{\#GROUND\_TRUTH\_HERE}}
    \item \textbf{Model Prediction}: {\color{red!70!black}\texttt{\#MODEL\_PREDICTION\_HERE}}
\end{enumerate}

\vspace{0.3cm}
Now, please provide your judgment in the specified JSON format.

\label{prompt_eval_judge}
\end{tcolorbox}

\begin{tcolorbox}[
    colback=white,
    colframe=black!75,
    fonttitle=\bfseries,
    title=Format SFT Prompt (Taking denoise as an example),
    listing options={basicstyle=\ttfamily\small,breaklines=true},
    breakable
]
\textbf{ROLE: Visual Reasoning Analyst}
\vspace{0.1cm}

You are a meticulous analyst responsible for generating detailed reasoning processes based on images, questions, and provided answers. Your task is to analyze visual content and provide step-by-step reasoning.

\vspace{0.3cm}
\textbf{CORE INSTRUCTIONS}
\vspace{0.1cm}

Your task is to:
\begin{itemize}
    \item Analyze the core of the problem, clarify what key information needs to be obtained from the picture.
    \item You are encourged to denoise the image to read the information during inference with the following API. 
    \item  def denoise(image: str, method: str = "median", kernel\_size: int = 3) \# Denoising filter (gaussian/median/bilateral) for restoring scanned or low-light images. Denoising method: gaussian $\|$ median $\|$ bilateral, default is median.
    \item  Try to change the method and kernel\_size to uncover the picture text. The API call must be enclosed within a $<$code$><$/code$>$ tag.
    \item  Complete reasoning based on the final adjusted picture to ensure logical coherence and accord with the provided answer
    \item  The reasoning process should show the analysis steps in detail, including the observation of the picture content, the extraction of key data and the logical derivation
\end{itemize}

\vspace{0.3cm}
\textbf{OUTPUT FORMAT}
\vspace{0.1cm}

You must return your analysis in the following format:
\small
\begin{verbatim}
<think>
First, I analyze the problem requirements: [analysis steps]. 
The key information needed includes: [specific details].
<code>
image_path = denoise(image_path, method="gaussian", kernel_size=[odd number])
</code>
<output><image></output>
After denoising, I observe: [observations]. The extracted data shows: [data points]. 
Logically, this leads to: [derivation process].
</think>
<answer>
[Concise final answer]
</answer>
\end{verbatim}
\normalsize

\vspace{0.3cm}
\textbf{Input Data:}
\vspace{0.1cm}

\begin{enumerate}
    \item \textbf{Question}: {\color{green!60!black}\texttt{\#QUESTION\_HERE}}
    \item \textbf{Answer}: {\color{orange!80!black}\texttt{\#ANSWER\_HERE}}
    \item \textbf{Image dimensions}: {\color{blue!70!black}\texttt{\#IMAGE\_SIZE\_HERE}}
\end{enumerate}

\vspace{0.3cm}
Now, please provide your analysis following the specified format.

\label{prompt_visual_reasoning}
\end{tcolorbox}


\end{document}